\documentclass{article}

\usepackage[preprint, nonatbib]{neurips_2020}

\usepackage[utf8]{inputenc} 
\usepackage[T1]{fontenc}    
\usepackage{hyperref}       
\usepackage{url}            
\usepackage{booktabs}       
\usepackage{amsfonts}       
\usepackage{nicefrac}       
\usepackage{microtype}      
\usepackage{amsmath}
\usepackage{amssymb}
\usepackage{tikz}
\usepackage{pgfplots}

\newcommand{\abs}[1]{\left\vert #1 \right\vert}

\DeclareMathOperator*{\argmin}{\arg\min}
\DeclareMathOperator{\local}{local}
\DeclareMathOperator{\E}{\mathbb{E}}

\title{Server Averaging for Federated Learning}

\author{George Pu \\
  NSF Center for Big Learning \\
  University of Florida \\
  \texttt{pu.george@ufl.edu} \\
  \And
  Yanlin Zhou \\
  NSF Center for Big Learning \\
  University of Florida \\
  \texttt{zhou.y@ufl.edu} \\
  \AND
  Dapeng Wu \\
  NSF Center for Big Learning \\
  University of Florida \\
  \texttt{dpwu@ufl.org} \\
  \And
  Xiaolin Li\\
  Cognization Lab\\
  Palo Alto, CA\\
  \texttt{xiaolinli@ieee.org}\\
}

\begin{document}

\maketitle

\begin{abstract}
Federated learning allows distributed devices to collectively train a model without sharing or disclosing the local dataset with a central server.
The global model is optimized by training and averaging the model parameters of all local participants.
However, the improved privacy of federated learning also introduces challenges including higher computation and communication costs.
In particular, federated learning converges slower than centralized training.
We propose the server averaging algorithm to accelerate convergence.
Sever averaging constructs the shared global model by periodically averaging a set of previous global models.
Our experiments indicate that server averaging not only converges faster, to a target accuracy, than federated averaging (FedAvg), but also reduces the computation costs on the client-level through epoch decay.

\end{abstract}

\section{Introduction}

Deep learning has emerged as a powerful tool for many industrial and scientific applications.
However, deep learning requires large centralized datasets, whose collection can be intrusive, for training.
The finalized model can either be deployed on a server or edge devices.
Federated learning circumvents this problem by shifting compute responsibility onto  clients, letting a central server aggregate the resulting artifacts.
In FedAvg, edge devices train on locally available data, while the server averages the finished models \cite{mcmahan2017communication}.
This is repeated for multiple rounds of communication.
The server never has possession of any potentially sensitive data.

When data is independently and identically distributed (IID), federated learning algorithms converge rapidly.
FedAvg takes as a few as 18 communication rounds to reach 99\% accuracy for 100 device federated MNIST \cite{mcmahan2017communication}.
When the client devices are statistically heterogeneous, learning a single global model becomes very difficult \cite{li2019federated}.
In such cases, it is more natural to learn personalized models.
Still, there are circumstances where a single global model is desired.
For example, different online businesses might want a model capable of flagging a wide spectrum of fraudulent schemes.
since fraudsters are often repeat offenders, scams attempted on one platform may be reused on others.

Techniques for faster federated learning on non-IID data range from the simple to the complex.
On the simple end, Momentum Federated Learning averages the momenta of different devices into a global momentum which is distributed at the start of each round \cite{liu2020accelerating}.
This enables clients to use momentum gradient descent as their optimizer, provably increasing the rate of convergence.
On the complex end, SCAFFOLD uses the gradient of the global model as a control variate to address drifting among client updates \cite{karimireddy2019scaffold}.
Notably these two methods double the amount of information submitted by devices to the server.

To deal with the communication and scalability challenges introduced by above methods, efforts has been made to reduce the amount of rounds required for server-client communication \cite{zhou2020distilled}.
FedPAQ has made an initial effort \cite{reisizadeh2020fedpaq} to periodically average and quantize the client models before making the server update.
Then, periodic averaging for both server and client models followed up quickly \cite{amiri2020federated}.

In this paper, we take a different approach, using server averaging to accelerate convergence.
We justify the technique using heuristic arguments and experimentally show that it reaches a given test accuracy faster than FedAvg.
Additionally, we propose decay epochs for reducing client computation while maintaining non-IID performance.

\section{Related Work}

The history of stochastic gradient methods dates back to 1951, and is usually mentioned as Robbins-Monro process \cite{robbins1951stochastic}.
One technique that has historically been used to improve SGD convergence is iterate averaging \cite{polyak1990new, polyak1992acceleration, ruppert1988efficient}, also often referred to as Polyak-Rupert averaging.
Recently, the stability of an averaging scheme that considers a non-uniform average of the iterates is discussed \cite{neu2018iterate}.
A weighted average is applied which decays in a geometric manner.
Neu et. al., show that the same regularizing effect can be done for SGD with the linear least-squares regression problem.

Federated learning techniques heavily rely on above mentioned averaging schemes.
FedAvg is the most popular aggregation method that averages parameters of local models element-wise.
There exists two major branches for improving FedAvg \cite{mcmahan2017communication}.
One is lead by FedProx \cite{sahu2018convergence} that applies a proximal term to the local lost function of each client and thresholding the local updates.
Another approach is to proposes different averaging schemes to either save the communication cost or to improve the performance \cite{wang2020federated}.

Safa et. al. explore iterate averaging in the context of block-cyclic SGD \cite{wu2019safa}.
Most federated learning algorithms assume that clients are chosen uniformly.
In practice, devices conduct local training only when idle, with devices falling into blocks according to their timezone.
More formally, we want to minimize
\begin{equation}
    \E_{z \sim \mathcal{D}} f(w, z) \text{ where } \mathcal{D} = \sum_{i=1}^{m} \mathcal{D}_i
\end{equation}
while sampling $n$ points from $\mathcal{D}_1, \dots, \mathcal{D}_m$ in order for $K$ cycles.
In this block-cyclic setting, SGD is worse by a factor of $\sqrt{mn/K}$.
However, learning personalized models for each block using Averaged SGD \cite{polyak1992acceleration}---taking the average of all SGD iterate as the final model parameters---provides the same performance guarantees as SGD with IID sampling.

Stochastic Weight Averaging (SWA) applies Averaged SGD to deep learning \cite{izmailov2018averaging}.
Izmailov et. al. note that SGD generally converges to points near the boundary of a wide flat region and that optima width has been conjectured to correlate with generalization.
The average of the SGD iterates then lies at the the center of this flat region.
Moreover, to ensure coverage of this flat region, SWA uses a cyclic or a high constant learning rate.
This algorithm has the benefit of low computational overhead---only the moving average needs to be recorded--and simplicity.
SWA does not improve the rate of convergence compared to SGD.
In fact, SWA converges to worse but better generalizing optima than SGD.


\section{Server Averaging}

Both papers \cite{karimireddy2019scaffold, liu2020accelerating} suggest that iterate averaging would be well suited for federated learning.
However, the time scales for client level averaging---10 to 20 epochs over a small dataset---are too low to provide any noticeable improvement over base FedAvg.
Instead, we perform server-level averaging of the global models.
Every $R$ communication rounds, the server averages $P$ previous global parameters to produce the next global model.
More precisely, after the $t \equiv 0 \pmod R$ round of FedAvg finishes,
\begin{equation}
    w^t \gets \sum_{i=0}^{P-1} w^{t-i}
\end{equation}
where $w^t$ are the model weights at round $t$.

This has three advantages.
\begin{enumerate}
    \item The global update step of FedAvg weakly resembles the update step for SGD.
    We denote the model parameters post-local training as $local(w^t, i)$, where $i$ is a numbered client.
    The FedAvg update for a random subset of $M$ clients $\mathcal{S}$ is 
    \begin{equation}
        \label{fedavg}
        w^{t+1} = \sum_{i \in \mathcal{S}} \frac{\abs{\mathcal{D}_i}}{\abs{\mathcal{D}}} \local(w^t, i).
    \end{equation}
    Let $\Delta w_i^t = w^t - \local(w^t, i)$. We can rewrite Equation~\ref{fedavg} as
    \begin{equation}
        w^{t+1} = w^t - \sum_{i \in \mathcal{S}} \frac{\abs{\mathcal{D}_i}}{\abs{\mathcal{D}}} \Delta w_i^t.
    \end{equation}
    FedAvg is an iterative optimization algorithm with learning rate 1.
    As such, it may be able to benefit from iterate averaging to accelerate convergence.
    \item Weight averaging can mitigate federated learning's struggles with heterogeneous data.
    Suppose the dataset $\mathcal{D}$ is poorly partitioned among $N$ client into $\mathcal{D}_1, \dots, \mathcal{D}_N$.
    Let $f(w, \cdot)$ be a non-convex loss function of the model weights $w$ over a set.
    Training on non-IID datasets converge to distant local optima $w^*_i = \argmin_w f(w, \mathcal{D}_i)$ which occupy a non-convex region of the loss function.
    Averaging these optima produces model parameters of high loss.
    \begin{equation}
        f\left( \frac{1}{N} \sum_{i=1}^{N} w_i, \right) > \max_{i=1,\dots,N} f(w^*_i, \mathcal{D})
    \end{equation}
    SWA shows that iterate averaging produces better generalizing optima than SGD alone.
    By learning better generalizing optima, they are more likely to lie in the same convex region.
    These server parameters are more easily trained by the clients.
    
    \item It has been noted that increasing the number of participating clients $M$ increases the effectiveness of federated learning \cite{mcmahan2017communication}.
    However, raising $M$ also increases the network load on the server and total communication cost of FedAvg.
    Additionally, clients are not always available for training, meaning that there is a hard upper bound on $M$.
    Averaging previous iterates allows the server to ``increase" $M$ without these downsides.
    Of course, because different rounds have different starting paramters, this is not actually equivalent to training more clients per round.
\end{enumerate}

We test server averaging on 100 client, non-IID federated MNIST.
Every round, 10 clients are chosen to participate with uniform probability: downloading, training, then uploading the model parameters.
The dataset is divided according to the procedure in Mcmahan et. al. \cite{mcmahan2017communication}.
First, sort the entire dataset by class label.
Break the dataset into shards of equal size $\abs{\mathcal{D}}/(2N)$ and assign 2 shards to each device.
For MNIST, this ensures clients have no more than 4 distinct classes (a shard may have 2 digits).
We use an initial learning rate of $\eta = 0.01$ and locally train for $E = 5$ epochs every communication round.
The number of communication rounds is capped at 500.

\begin{figure}[t]
    \centering
    \includegraphics[width=\linewidth]{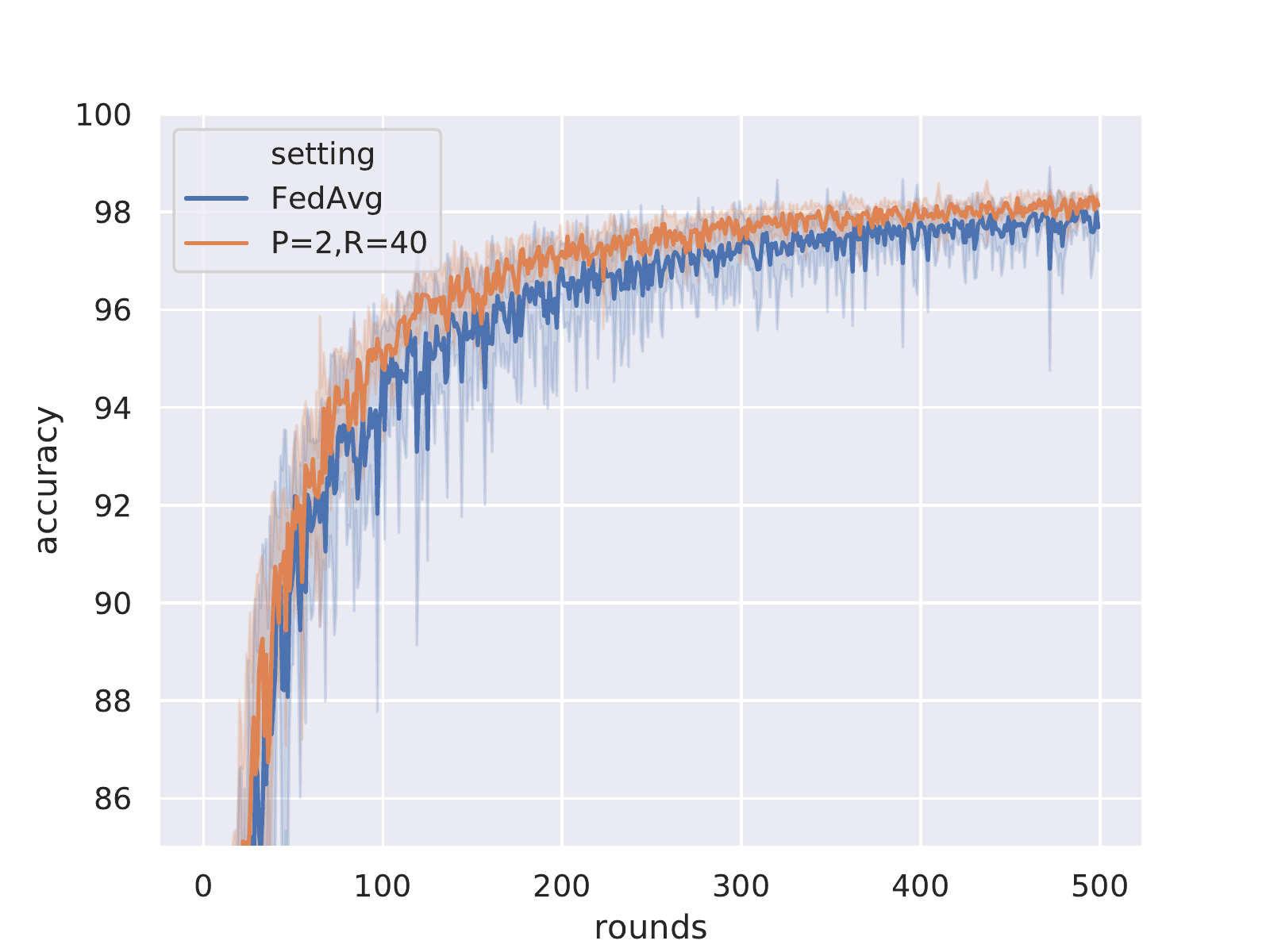}
    \caption{Comparison plot between FedAvg and server averaging ($P=2,R=40$).}
    \label{fig:server_avg}
\end{figure}

Let $T_a$ be the number of communication rounds required to reach $\geq a$ test accuracy.
We report the mean and standard deviation as $mean \pm std$ for $T_{90}, T_{95}, T_{97}, T_{98}$ over 5 trials in Table~\ref{tab:server_avg}.
The lowest value in each column are bolded.
Notably, server averaging universally performs better than plain FedAvg, achieving the accuracy thresholds faster.
Only three cells in Table~\ref{tab:server_avg} have a larger rounds to test accuracy.
The average improvement over all $P, R$ values is $1.98, 8.22, 27.04, 18.67$ for $T_{90}, T_{95}, T_{97}, T_{98}$ compared to FedAvg.
Surprisingly, the gains for $T_{97}$ are greater than $T_{98}$.
The average test accuracy over 5 trials for two hyperparameter settings are given in Figure~\ref{fig:server_avg}.

\begin{table}[t]
    \centering
    \caption{Mean round and standard deviation to $\{90, 95, 97, 98\}$\% accuracy for non-IID federated MNIST. The first row are baseline results for FedAvg.}
    \label{tab:server_avg}
    \begin{tabular}{c c l l l l}
        \toprule
        $P$ & $R$ & $T_{90}$ & $T_{95}$ & $T_{97}$ & $T_{98}$ \\
        \midrule
        1 & 1 & 36.40 $\pm$ 10.04 & 85.40 $\pm$ 20.12 & 179.40 $\pm$ 42.51 & 309.40 $\pm$ 27.22 \\
        \midrule
        2  & 10 & 31.60 $\pm$ 5.27 & 75.40 $\pm$ 17.78 & 152.80 $\pm$ 18.07 & 302.40 $\pm$ 4.83 \\
        3  & 10 & 31.20 $\pm$ 4.71 & 69.40 $\pm$ 7.20 & 144.80 $\pm$ 11.97 & 288.80 $\pm$ 13.46 \\
        4  & 10 & 32.00 $\pm$ 4.36 & 71.80 $\pm$ 7.66 & 147.60 $\pm$ 17.92 & 296.60 $\pm$ 15.84 \\
        5  & 10 & 42.80 $\pm$ 9.07 & 89.80 $\pm$ 18.21 & 165.20 $\pm$ 35.92 & 294.80 $\pm$ 36.49 \\
        \midrule
        2  & 20 & 35.80 $\pm$ 16.51 & 96.60 $\pm$ 41.95 & 172.60 $\pm$ 65.19 & 300.40 $\pm$ 31.45 \\
        3  & 20 & 31.40 $\pm$ 4.83 & 71.40 $\pm$ 5.59 & 148.00 $\pm$ 16.08 & 293.80 $\pm$ 15.99 \\
        4  & 20 & 31.60 $\pm$ 5.77 & 73.40 $\pm$ 7.64 & 154.60 $\pm$ 16.38 & 309.60 $\pm$ 15.58 \\
        5  & 20 & 35.60 $\pm$ 3.78 & 84.40 $\pm$ 19.87 & 160.60 $\pm$ 33.10 & 299.40 $\pm$ 27.74 \\
        \midrule
        2  & 40 & 28.00 $\pm$ 5.39 & 65.60 $\pm$ 5.73 & 127.80 $\pm$ 15.58 & 263.80 $\pm$ 18.78 \\
        3  & 40 & 37.80 $\pm$ 10.47 & 76.20 $\pm$ 11.69 & 158.20 $\pm$ 30.42 & 296.60 $\pm$ 29.92 \\
        4  & 40 & 36.40 $\pm$ 10.21 & 80.80 $\pm$ 19.02 & 155.40 $\pm$ 41.34 & 278.40 $\pm$ 30.45 \\
        5  & 40 & 36.00 $\pm$ 11.73 & 69.60 $\pm$ 7.89 & 141.20 $\pm$ 26.50 & 275.80 $\pm$ 30.14 \\
        \bottomrule
    \end{tabular}
\end{table}

\section{Decay Epochs}

Continuing our investigation of sever level strategies for accelerating convergence, we explore epoch decay.
Epoch decay is analogous to learning rate decay for SGD.
Every $D$ communication rounds, the number of local epochs $E_t$ is reduced in half from a starting value $E$.
The number of local epochs cannot be less than 1.
\begin{equation}
    E_t = \max\left( E \frac{1}{2^{\lfloor t / D \rfloor}}, 1 \right)
\end{equation}
Epoch decay reduces the distance traveled during local training by limiting the number of updates.
As demonstrated by SWA \cite{izmailov2018averaging}, advanced SGD training tends towards the boundary of a low loss region.
With regard to the test loss, much of this movement has little impact.
It is also known that setting the local epochs too large hurt convergence \cite{mcmahan2017communication}.
By decaying the epochs, we can avoid this situation, especially in the delicate final phase of federated learning.

We test different epoch decay rates on 100 client non-IID MNIST.
The results are reported in Table~\ref{tab:epoch_decay}.
Here the results are not as strong.
The average improvement over all decay epochs $D$ is $2.5, 13.07, 27.27, 7.53$ for $T_{90}, T_{95}, T_{97}, T_{98}$ compared to FedAvg.
However, it is still notable given that the amount of computation is reduced by up to 40\%.
There is little correlation between the computation reduction and $T_a$ values.
The value of computation is not constant throughout a federated learning session but decays over time.
In fact, it may be possible to start with a high number of local epochs (e.g. $E = 20$) then aggressively decay this amount as the communication rounds pass.

\begin{table}[t]
    \centering
    \caption{Mean round and standard deviation to $\{90, 95, 97, 98\}$\% accuracy for non-IID federated MNIST}
    \label{tab:epoch_decay}
    \begin{tabular}{c l l l l}
        \toprule
        $D$ & $T_{90}$ & $T_{95}$ & $T_{97}$ & $T_{98}$ \\
        \midrule
        - & 36.40 $\pm$ 10.04 & 85.40 $\pm$ 20.12 & 179.40 $\pm$ 42.51 & 309.40 $\pm$ 27.22 \\
        \midrule
        100 & 30.00 $\pm$ 6.04 & 64.80 $\pm$ 6.22 & 138.80 $\pm$ 19.94 & 296.80 $\pm$ 14.34 \\
        125 & 47.80 $\pm$ 35.42 & 79.00 $\pm$ 16.97 & 169.60 $\pm$ 23.61 & 316.80 $\pm$ 13.97 \\
        150 & 31.80 $\pm$ 3.49 & 71.20 $\pm$ 6.22 & 150.80 $\pm$ 31.40 & 307.20 $\pm$ 24.87 \\
        200 & 27.20 $\pm$ 17.53 & 73.20 $\pm$ 11.84 & 153.20 $\pm$ 27.22 & 286.00 $\pm$ 36.45 \\
        225 & 37.00 $\pm$ 2.55 & 75.40 $\pm$ 4.93 & 150.60 $\pm$ 13.65 & 302.00 $\pm$ 19.54 \\
        250 & 29.60 $\pm$ 4.22 & 70.40 $\pm$ 5.18 & 149.80 $\pm$ 10.43 & 302.40 $\pm$ 20.31 \\
        \bottomrule
    \end{tabular}
\end{table}

\section{Conclusion}
This paper improves upon an existing federated learning algorithm by performing  periodic server-side averaging.
The proposed adaptation of FedAvg has three major benefits: (1) it uses iterate averaging for accelerated convergence, (2) it learns a better generalizing optima than SGD, (3) the effectiveness of FL is increased due to recycling of previously participating clients.
We empirically show that server averaging takes fewer rounds than FedAvg to a desired accuracy level.
In addition, we propose epoch decay to lower the computation costs for each client.
Epoch decay limits the number of updates, similar to learning rate decay for SGD, and reduces the amount of computation by up to 40\%. 

In the future, we wish to extend the server averaging to both various neural network types (i.e. attention, LSTM, etc.) and layer-wise building blocks (i.e. batch normalization layers, etc.).
In addition, we wish to investigate the performance of epoch decay paired with state-of-the-art update methods such as match averaging \cite{wang2020federated}.

\newpage
\bibliographystyle{plain}
\bibliography{refs}

\end{document}